\documentclass[sigconf]{acmart}

\usepackage{enumitem}
 \usepackage{bm}
 \usepackage{amsmath}
\usepackage{svg}
\usepackage{arydshln}
\usepackage{multirow}
\usepackage{geometry}
\usepackage{arydshln}
\usepackage{graphics}
\usepackage{caption}
\usepackage{threeparttable}
\usepackage{graphicx}
\usepackage{wrapfig}
\usepackage{xcolor}  
\usepackage{tabularx}
\usepackage{booktabs,multirow,diagbox}
\usepackage{float}
\usepackage{subfig}
\usepackage{algorithm,algorithmic}
\usepackage{stmaryrd}
\definecolor{ggray}{HTML}{E7E6E6}
\usepackage{colortbl}
\usepackage{balance}
\newcommand{\hide}[1]{} 

\newcommand{\vpara}[1]{\vspace{0.05in}\noindent\textbf{#1 }}

\copyrightyear{2025} 
\acmYear{2025} 
\setcopyright{acmlicensed}
\acmConference[KDD '25]{Proceedings of the 31st ACM SIGKDD Conference on Knowledge Discovery and Data Mining V.1}{August 3--7, 2025}{Toronto, ON, Canada}
\acmBooktitle{Proceedings of the 31st ACM SIGKDD Conference on Knowledge Discovery and Data Mining V.1 (KDD '25), August 3--7, 2025, Toronto, ON, Canada}
\acmDOI{10.1145/3690624.3709320}
\acmISBN{979-8-4007-1245-6/25/08}

\begin{document}

\title{How to Use Graph Data in the Wild to Help\\ Graph Anomaly Detection? }


\author{Yuxuan Cao}
\affiliation{
  \institution{Zhejiang University}
  \institution{Fudan University}
  \country{}
}
\authornote{This work was done when the author was a visiting student at Fudan University.}
\email{caoyx@zju.edu.cn}

\author{Jiarong Xu}
\affiliation{%
  \institution{Fudan University}
   \country{}
}
\authornote{Corresponding author.}
 \email{ jiarongxu@fudan.edu.cn}
 
\author{Chen Zhao}
\affiliation{%
  \institution{Fudan University}
  \country{}
}
 \email{czhao23@m.fudan.edu.cn}
 
\author{Jiaan Wang}
\affiliation{%
  \institution{WeChat AI, Tencent}
  \country{}
}
 \email{jawang.nlp@gmail.com}

\author{Carl Yang}
\affiliation{%
  \institution{Emory University}
  \country{}
}
\email{j.carlyang@emory.edu}

\author{Chunping Wang}
\affiliation{%
  \institution{Finvolution Group}
  \country{}
}
 \email{ wangchunping02@xinye.com}

 \author{Yang Yang}
\affiliation{%
  \institution{Zhejiang University}
  \country{}
  }
 \email{ yangya@zju.edu.cn}

\renewcommand{\shortauthors}{Yuxuan Cao et al.}
\begin{abstract}
In recent years, graph anomaly detection has gained considerable attention and has found extensive applications in various domains such as social, financial, and communication networks. 
However, anomalies in graph-structured data present unique challenges, including \emph{label scarcity}, \emph{ill-defined anomalies}, and \emph{varying anomaly types}, making supervised or semi-supervised methods unreliable. Researchers often adopt unsupervised approaches to address these challenges, assuming that anomalies deviate significantly from the normal data distribution. Yet, when the available data is insufficient, capturing the normal distribution accurately and comprehensively becomes difficult. To overcome this limitation, we propose to utilize external graph data (\emph{i.e.}, graph data in the wild ) to help anomaly detection tasks. This naturally raises the question: \emph{How can we use external data to help graph anomaly detection task?} 
To answer this question, we propose a novel framework \emph{Wild-GAD}. Our framework is built upon a unified database, UniWildGraph, which comprises a large and diverse collection of graph data with broad domain coverage, ample data volume, and a unified feature space. Further, we develop selection criteria based on \emph{representativity} and \emph{diversity} to identify the most suitable external data for each anomaly detection task.
Extensive experiments on six real-world test datasets demonstrate the effectiveness of Wild-GAD. Compared to the baseline methods, our framework  has an average 18\% AUCROC
and 32\% AUCPR improvement over the best-competing
methods.
\end{abstract}

\maketitle
\section{INTRODUCTION} \label{sec:intro}
Graphs are ubiquitous in the real world and commonly utilized to represent the complex inter relationships among objects across a wide range of domains, such as social networks~\cite{davies2022realistic}, financial networks~\cite{chai,huang2022dgraph}, biological networks~\cite{zhang2022ssgnn}, and traffic networks~\cite{yu2017spatio}. 
Graph Anomaly Detection (GAD) focuses on identifying abnormal instances or outliers that deviate from the prevailing patterns of normal nodes within a graph. This task holds significant practical value in numerous applications, including detecting malicious activities in social media, identifying fraudulent transactions in financial systems, and uncovering fraudsters in telecommunications networks.
The growing adoption of Graph Neural Networks (GNNs) has spurred substantial progress in GAD. Supervised and semi-supervised GAD methods~\cite{BernNet,AMNet,BWGNN,GATSep,DCI,GHRN} leverage labeled anomaly data as supervision, enabling the learning of both normal and abnormal distributions to distinguish anomalous nodes from normal ones effectively. 

However, anomalies in graph-structured data exhibit several unique characteristics that pose significant challenges: 
(1) \emph{{label scarcity}}~\cite{ma2023comprehensive,ma2024graph}: The number of labeled anomalies is typically limited, as real-world scenarios often make it costly and time-consuming to perform double-checking and cross-validation during the manual labeling process by experts.
(2) \emph{{ill-defined anomalies}}~\cite{foorthuis2021nature}: Anomalies in graphs often lack explicit and accurate definitions. For instance, a credit company may label users with more than ten late payments as anomalous. However, users with late payments close to this threshold (\emph{e.g.}, eight or nine instances) may still exhibit anomalous behavior, illustrating the ambiguity in defining anomalies.
(3) \emph{{varying types}}~\cite{wang2019detecting}: The behaviors of anomalies can evolve over time, introducing new, unseen abnormal patterns. This dynamic nature makes it challenging to accurately learn and generalize the abnormal distributions.
As a result, supervised and semi-supervised methods that rely on limited, ill-defined, and dynamically changing abnormal labels often fail to  capture the full range of anomaly patterns, limiting their reliability and effectiveness. 

To overcome these limitations, many works have proposed employing unsupervised strategies~\cite{DOMINANT,AnomalyDAE,AdONE,GAAN,CoLA,CONAD,GAE,OCGNN}, focusing on learning a comprehensive and robust distribution of normal patterns rather than abnormal patterns. Based on well-learned normal patterns, anomalies that significantly deviate from these patterns can be naturally distinguished. 
However, these methods heavily depend on the presence of sufficient and representative normal patterns in the target data, which may not always be available. When normal data is insufficient, the resulting normal distribution can be inaccurate, thereby undermining the performance of anomaly detection.
For example, in financial networks, most users engage in high-frequency transactions, causing models to predominantly learn these frequent behaviors while overlooking low-frequency ones. This bias can lead to the misclassification of legitimate low-frequency traders as anomalies. However, true anomalies in this context are not simply defined by low frequency but by distinct irregularities, such as irregular intervals or interactions with uncommon counterparties. 

To effectively model the distribution of normal patterns and identify anomalies as deviations from these patterns, it is essential to provide the unsupervised model with a more comprehensive view of normal behaviors. It enables the model to capture normal distributions more accurately, thereby reducing the risk of misclassifying legitimate but underrepresented behaviors as anomalies.
In this work, we propose borrowing knowledge from external data (\emph{i.e.}, graph data in the wild) to address such an issue. 
This leads to a natural question:  \emph{How can we use external data to help graph anomaly detection tasks?} 

To answer this question, we design a framework called Wild-GAD. It primarily consists of three components: a diverse database to ensure the model have access to a wide range of representative data resources, a data selection strategy to identify the most useful external graph data, and a training strategy that utilizes the selected external data to effectively perform the anomaly detection task. Putting together,  these components work cohesively to enhance graph anomaly detection by wisely leveraging external knowledge.

However, implementing this framework presents several challenges. The first challenge is the availability and applicability of external graph data in the wild.  Currently, no existing library or database can be directly applied to this task. The ideal database must meet three key requirements: (1) \emph{Various domain coverage}: It should cover a wide range of domains, ensuring that target networks from any domain can benefit from it.
(2) \emph{Sufficient amount and applicability}: It must contain a large amount of external data, ensuring the model has enough samples to learn normal patterns effectively.
(3) \emph{Unified feature space}: External graph data often originates from various sources, and the features can differ significantly in terms of both dimensions and semantic meanings.
A unified feature space is essential to process and integrate diverse feature types.  
To tackle this challenge, we propose a database called {UniWildGraph}. It includes twelve external graph datasets from four distinct domains as raw data to ensure domain diversity. To enhance the database and better adapt it for downstream tasks, we apply four types of augmentations with varying augmentation ratios. Finally, to standardize all features into a consistent textual format, we leverage language models to align them within a unified semantic space.

With the constructed database {UniWildGraph} containing a vast amount of external graph data, not all of them necessarily have a positive impact on improving the anomaly detection task on the target graph. External graphs may not share relevant characteristics with the target data, and including irrelevant or redundant graphs could introduce noise. Therefore, it is essential to design a selection strategy that identifies the most useful external graphs while operating within the constraints of a limited resource budget. To tackle this challenge, we propose a selection strategy based on two key criteria: \emph{representativity} and \emph{diversity}. The representativity criterion ensures that the selected data is sufficiently relevant to the target data.
On the other hand, the diversity criterion ensures that the selected external graphs offer a broad and comprehensive coverage of normal patterns, enabling the unsupervised model to learn more generalizable and robust representations of normal behavior. 

Given the selected external data, we take the model first trained on the target data as the initialization, and then train the model on the selected data.  This training process enables the model to incorporate the diverse and comprehensive normal patterns provided by the external data while retaining the knowledge learned from the target data.  
Consequently, the model develops a more accurate and generalized understanding of the normal distribution, so that anomalies are identified as samples that significantly deviate from this learned normal distribution. In this way, the model's ability to effectively and reliably distinguish abnormal patterns is significantly enhanced.
Our key contributions are as follows:
\vspace{-0.1in}
\begin{itemize}
    \item To the best of our knowledge, we are the first to identify the necessity and leverage the knowledge of external graph data to help graph anomaly detection task. We propose a framework called Wild-GAD that allows users to select and adapt the most suited external graph data for their target data to build a better graph anomaly detection model. 
    
    \item We construct a database called UniWildGraph comprising a massive amount of graph data with the advantages of \emph{various domain coverage}, \emph{sufficient amount and flexible to various detection tasks} and \emph{unified feature space}. 
    Users with target graph from any domain can leverage data from  UniWildGraph to help anomaly detection task.


\item We introduce a selection strategy based on \emph{representativity} and \emph{diversity} criteria to  choose the most effective external graph data for anomaly detection task on target graph.

\item 
To evaluate the effectiveness of Wild-GAD, we tested it on six real-world datasets. Wild-GAD outperformed baseline methods, achieving an average improvement of 18\% in AUC-ROC and 32\% in AUC-PR over the best competing methods. 
Compared to semi-supervised methods, Wild-GAD can surpass competitors at label ratios of up to 70\%.

\end{itemize}

\section{RELATED WORK} \label{sec:related}
\vpara{Graph Anomaly Detection.} 
With the development of GNNs, many GAD studies focus on designing GNN algorithms~\cite{AnomalyDAE,GAE, BernNet, AMNet,BWGNN,GATSep,DCI,GHRN}.
Due to the high cost and unavailability of manually labeling the ground truth anomalies, existing efforts are mostly unsupervised~\cite{DOMINANT, AnomalyDAE, OCGNN, CoLA, CONAD}.  DOMINANT~\cite{DOMINANT} uses a GCN-based autoencoder framework to compress and reconstruct the initial graph. The reconstruction error for each node can be used to identify anomalies. 
OCGNN~\cite{OCGNN} combines the powerful representation ability of GNNs with the classical hypersphere learning objective to detect anomalies.
Another line aims to capture the normal patterns in the graph through a self-supervised learning task. CoLA~\cite{CoLA} conducts contrastive instance pair sampling between each node and its neighboring and uses an anomaly-related objective to train the contrastive learning model. CONAD~\cite{CONAD} leverages prior human knowledge of anomaly types as contrastive samples and integrates them into the detection model through a contrastive loss.
Apart from basic paradigms, TAM~\cite{TAM} identifies the one-class homophily in GAD datasets with both injected and real anomalies, and leverages this property to propose the local node affinity as novel unsupervised GAD measure. RAND~\cite{RAND} uses reinforcement learning to expand and select candidate neighborhoods, while enhancing anomaly detection with a consistent message aggregator. 
However, these approaches merely focus on their own network, which may lack robustness and generalizability.

A somewhat related line of research is cross-domain graph anomaly detection, which uses cross-domain data to improve performance within a specific domain~\cite{xu2023metagad,wang2023cross}.
MetaGAD~\cite{xu2023metagad}  designs a meta-learning method to transfer
the knowledge between unlabeled and labeled nodes for GAD.
It leverages synthetic anomalies to improve model performance on target GAD.
ACT~\cite{wang2023cross}  aligns the representations of the target graph with those of labeled normal nodes in the source graph.
It also ensures that the representations of normal nodes in the source graph deviate significantly from those of anomalous nodes.
However, these methods require a large number of high-quality anomaly labels and closely aligned source data with identical labels and feature spaces, limiting their practicality in real-world scenarios.
Different from these works, our work does not require anomaly labels and aim  to leverage complementary external knowledge of graph data in the wild to help model the normal distribution well in an unsupervised manner.

\vpara{External Knowledge Exposure for Anomaly Detection.}
In the field of CV and NLP,  several approaches for anomaly detection have leveraged the knowledge from pre-trained models to discern normal data patterns. Among these methods are DN2~\cite{DN2}, PANDA~\cite{PANDA}, and MSAD~\cite{MSAD}. They initially harness pre-trained models to extract features characterizing standard data instances. Subsequently, they apply techniques such as k-nearest neighbors (KNN) and Gaussian mixture models (GMM) to measure the deviation of new data points from the identified normal feature set, using this distance to compute an anomaly score. However, in the field of graph anomaly detection, there is still no works focus on leveraging external knowledge to help promote the model.
Moreover, outlier exposure (OE) ~\cite{OE}, is also an innovative technique for anomaly detection that employs an auxiliary dataset of outliers. However, OE faces a significant challenge: it necessitates a large and varied set of outlier data for training, which is often unavailable in real-world applications. Furthermore, the learned representations may not effectively generalize to previously unseen outlier distributions, and the presence of irrelevant outliers can negatively impact performance.

\section{BASIC ONE-CLASS  GRAPH ANOMALY DETECTION MODEL}
This section reviews the one-class assumption-based graph anomaly detection model commonly used in related literature. Our work follows Deep One-Class SVDD~\cite{ruff2018deep} paradigm as the backbone model. 
 One-Class SVDD model focuses on describing the behavior of dominated normal data by finding a hypersphere with a minimal radius in the embedding space. 
The basic assumption is that most normal data share similar behavior patterns and can be located within a hypersphere in the embedding space. While those anomalous do not conform to the expected normal behavior and their positions deviate far from the center of the hypersphere. So that the optimal hypersphere can serve as the distinguishing rule for detecting anomalies.
Specifically in the field of graph learning, we denote an attributed graph $\mathcal{G}=(\mathcal{V}, A, X)$, where $\mathcal{V}$ is the node set, $N=|\mathcal{V}|$ is the number of nodes, $A \in\{0,1\}^{N \times N}$ is the adjacency matrix. $X \in \mathbb{R}^{N \times d_{i n}}$ is the input node feature matrix and $d_{in}$ indicates the dimension of input features.  An attributed graph is first fed into a graph neural network $\phi_{\mathcal{W}}:\mathbb{R}^{d_{in}} \rightarrow \mathcal{S}^{d}$, and get corresponding embeddings $Z$, where $\mathcal{W} = \left\{\boldsymbol{W}^1, \ldots, \boldsymbol{W}^L\right\}$ denotes the weights of the graph neural network $\phi$ and $d$ is the embedding dimension. The model aims to construct a hypersphere in the embedding space to gather the most normal nodes around the center $\mathbf{c}$ of the hypersphere. This can be done by minimizing the following loss:
\begin{equation}
    \mathcal{L} = \frac{1}{\beta N} \sum_{i=1}^N\left[\left\|\phi(\boldsymbol{X}, \boldsymbol{A} ; \mathcal{W})_i-\boldsymbol{c}\right\|^2-r^2\right]+r^2+\frac{\lambda}{2} \sum_{l=1}^L\left\|\mathbf{W}_l\right\|^2,
\end{equation}
where $r$ is the radius of the hypersphere, $\beta \in[0,1)$ is an upper bound on the fraction of training errors and a lower bound of the fraction of support vectors, $\boldsymbol{c}$ is the center of the hypersphere defined as the average of the embeddings of the nodes.  The third term indicates the weight decay regularization with hyper-parameter $\lambda$. 

Accordingly, each node  the anomaly score is measured by the distance between the center and the embedding $z_i$ of each node $v_i$.
\begin{equation}
    s_i=\left\|z_i-\boldsymbol{c}\right\|^2 .
\end{equation}

It should be emphasized that, unlike previous research on One-Class Graph Neural Networks (OCGNN)~\cite{OCGNN}, which solely focuses on normal data during the training process and can be regarded as a semi-supervised learning approach, our method operates in an unsupervised manner.

\section{METHODOLOGY}\label{sec:model}
   \begin{figure*}[t]
    \centering
    \includegraphics[width=1.0\textwidth]{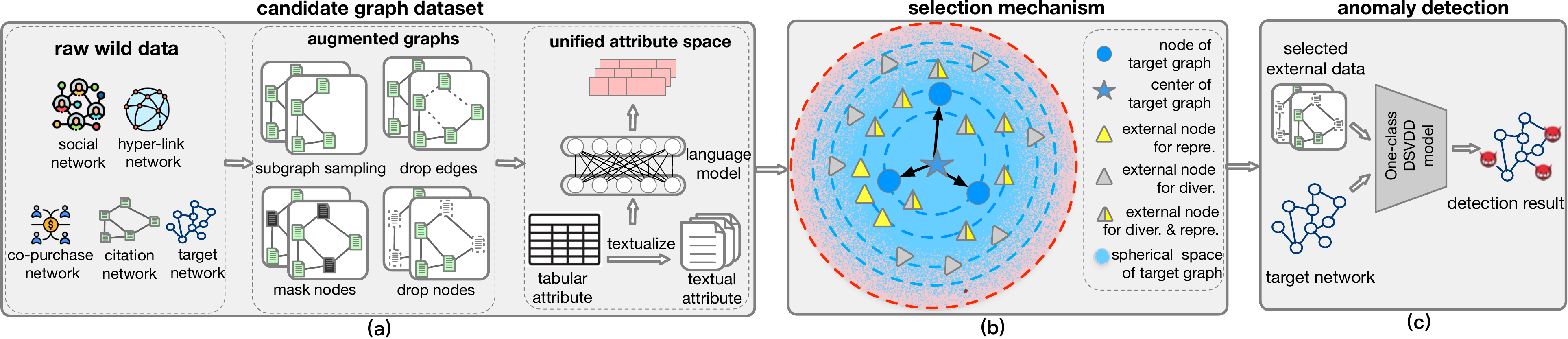}
    \captionsetup{font={small}}
    \caption{Overview of the proposed Wild-GAD framework. 
    } 
     \label{fig: framework}
\end{figure*}
In this section, we first introduce the proposed Wild-GAD framework, which aims to wisely leverage unlabeled wild data to help promote graph anomaly detection. The overall pipeline is illustrated in Figure~\ref{fig: framework}. The Wild-GAD framework consists of three major components, the constructed candidate datasets, selection criteria, and training strategy. 
The rest of this section is organized as follows. We further describe our collection and preparation of raw wild graph data in Section~\ref{subsec:dataset}. Then in Section~\ref{subsec:select},  we present the selection criteria of wild data. Finally, we propose the overall training strategy in Section~\ref{subsec:training}.

\subsection{Candidate Dataset Construction}\label{subsec:dataset}
To build a well-prepared candidate wild database, we focus on providing a broad and diverse range of external data, which is crucial for accurately and comprehensively modeling normal behavior. As illustrated in Figure~\ref{fig: framework}(a), the dataset is designed to meet three key characteristics outlined in the introduction: \emph{varied domain coverage}, \emph{sufficient quantity and relevance}, and a \emph{unified feature space}.
The following elaborates on each characteristic and describes our approach to incorporating them into the dataset construction.

\vpara{Rough data in the wild for graph anomaly detection.}
We consider the diversity of raw data from two perspectives: different domains and varying scales.
To ensure a comprehensive representation of normal patterns across domains, we have gathered raw graph data from the wild within four distinct domains: \emph{E-Commerce}, \emph{Social Network}, \emph{Citation}, and \emph{Hyper-Link}, as shown in the left part of Figure~\ref{fig: framework} (a). Across these domains, we have assembled twelve datasets, including P-Learning, P-Electronic, P-Entertainment, P-Household, P-Fashion, YelpRes, YelpNYC, Instagram, Reddit, Cora, PubMed, arXiv, and WikiCS. When target data is available, it can be included to provide a more accurate and realistic reference. 
Additionally, our datasets span a wide range of graph sizes to capture diverse graph structures. For example, the smallest graph, Cora, contains 2,708 nodes and 10,858 edges, while the largest, P-Learning, has 651,762 nodes and 24,599,888 edges. This variety in scale improves the framework's capacity to identify and characterize normal behaviors across different graph sizes, enhancing its ability to capture normal patterns effectively. More detailed descriptions of the characteristics of raw datasets can be found in Appendix~\ref{app:dataset}.

\vpara{Augmentations can serve as good candidates.}  In addition to the original graphs collected in Section~\ref{subsec:dataset}, we incorporate graph data augmentations to diversify our candidate datasets, thereby achieving broader coverage of normal patterns. 
Previous studies have highlighted that certain augmentations  (\emph{e.g., subgraph extraction}~\cite{Zhao2021DataAF,you2020graph,zhu2021graph}, \emph{node dropping}~\cite{you2020graph,Zhao2021DataAF}) can reduce redundancy in the original data, making them more suitable for downstream tasks~\cite{Zhao2021DataAF} (\emph{e.g.}, community detection, anomaly detection). 
For instance, in email communication networks, frequent correspondences between managers and secretaries are considered normal patterns, characterized by high connectivity typical of hub structures.
To capture such task-relevant information, we can employ subgraph sampling augmentation, which focuses on high-connectivity substructures while reducing the need to analyze the entire graph.
Furthermore, augmentations have been shown to enhance the generalizability and robustness of models~\cite{Han2022GMixupGD}, providing particular advantages in cross-dataset scenarios by enabling better adaptability to diverse data distributions.
Following common practices in graph augmentation, as illustrated in the middle part of Figure~\ref{fig: framework}(a), we adopt four augmentation operations, denoted as $\mathcal{A}$:
\begin{itemize}
    \item \textbf{Feature masking} sets certain node feature values to zero or the average of their neighbors;
    \item \textbf{Node dropping} removes a certain portion of nodes from the original graph;
    \item \textbf{Edge perturbation} involves modifying the original graph's structure by adding or removing edges, resulting in a new graph with altered connectivity;
    \item \textbf{Subgraph Extraction} extracts a subset of nodes along with their edges and features to form a subgraph.
\end{itemize}
To control the degree of augmentation applied to the original graph, we apply each type of augmentation with predefined ratios $\tau \in {0.2, 0.4, 0.6, 0.8}$.
Given an original graph $G$, we define the set of augmented graphs as $\mathcal{A}_G:=\left\{a_r(g) \mid a \in \mathcal{A} \wedge \tau \in \left\{0.2, 0.4, 0.6, 0.8\right\}\right\}$, where $a_\tau$ represents a transformation with ratio $\tau$. As detailed in Table~\ref{tab:external_data}, by integrating all original graphs with their augmented versions, the candidate database contains a total of 221 graphs.

\vpara{Feature Space Alignment.}
 Since the collected graph data in the wild come from diverse domains and sources, their features typically exist in different spaces and exhibit intricate characteristics. 
Motivated by the success of Language Models (LMs), as shown in the right part of Figure~\ref{fig: framework}(a), we propose transforming these heterogeneous attributes into textual representations and leveraging LMs to comprehend them within a unified space.

We categorize the features in graph data into two primary types: textual features and tabular features. 
Textual features are inherently textual in nature, such as those found in citation networks and Wikipedia networks. These features can be directly utilized as inputs to Language Models (LMs) without any transformation.

Tabular features are typically represented as vectors in categorical or numerical formats.  For instance, in social networks, node features often correspond to user profiles (\emph{e.g.}, age, gender and salary). 
 To process these tabular features, they must first be converted into meaningful textual representations. Following~\cite{tan2024walklm}, we design a rule-based method $\mathcal{P}$ for this transformation into text.
For example, consider a worker $v_i$ from the Tolokers Dataset with tabular features \emph{approved\_rate:} 0.8, \emph{skipped\_rate:} 0.2, \emph{expired\_rate:} 0.2, \emph{rejected\_rate:} 0.1. 
Our method converts this data into the textual representation: $\mathcal{P}(v_i)=$ \emph{"A worker whose approval rate is 0.8, skipped rate is 0.2, expired rate is 0.2, and rejection rate is 0.1."}.
  
After converting raw tabular features into text, we leverage the extensive language understanding capabilities of LMs to align the semantic spaces of heterogeneous features.  To handle the multilingual nature of our broad candidate datasets (e.g., English and Chinese), we employ a multilingual pre-trained NLP model, specifically mBERT~\cite{wu2019beto}, to ensure effective processing across diverse languages.

\subsection{Data Selection}\label{subsec:select}
As previously discussed in section~\ref{sec:related}, {the detection results of the one-class SVDD model depend on the expected standard hypersphere assumption. } However, insufficient and low-quality input data can lead to an irregular embedding space that does not adhere to the standard hypersphere, resulting in errors during hypersphere evaluation. To address this issue, we propose a strategy to selectively incorporate external data that can aid in model training and help fill out the embedding space, thereby providing a more comprehensive representation of normal patterns. In this section, we formally define the spherical embedding space and present the data selection strategy for choosing appropriate external data.

\vpara{Spherical Embedding Space.}
To achieve a more comprehensive and discriminative characterization of normal samples, we propose to analyze the learned embeddings within the spherical space. 
This spherical space is defined based on three key components:

\begin{itemize}
\item \textbf{Spherical center}: Represented as the blue star in Figure~\ref{fig: framework}(b), this center is calculated as the average embedding of all nodes, $\mathbf{c} = \frac{1}{n} \sum_{i=1}^n \boldsymbol{z}_i$.
\item \textbf{Radial distance}: The distance between each node (\emph{i.e.}, the blue points in Figure~\ref{fig: framework}(b)) and the spherical center, given by  $r_i=|| \boldsymbol {z}_i-\boldsymbol{c}||$.
\item \textbf{Relative direction}: The directional vector from each node to the spherical center, computed as $\boldsymbol{\varphi_i} = \left(\boldsymbol{z}_i-\boldsymbol{c}\right) /\left\|\boldsymbol{z}_i-\boldsymbol{c}\right\|$.
\end{itemize}
Thus, by simultaneously measuring from both angular and radial perspectives, each $d-$dimensional node embedding on the hyperspherical surface can be effectively represented using spherical coordinates, defined as $\boldsymbol{\rho_i}=[r_i,\boldsymbol{\varphi_i}]$. 

To apply this concept in practice, consider the target graph $\mathcal{G}_t$. 
We first train a base model and get the optimal parameter $\phi_0$ on it and obtain the corresponding learned spherical space.
This process yields the node embeddings $\boldsymbol{z}_i$ and spherical coordinates $\boldsymbol{\rho_i^t}$ for each node $v_i$ in $\mathcal{G}_t$. 
The base model $\phi_0$ encapsulates the current understanding of normal patterns within $\mathcal{G}_t$.
Building on this foundation, we apply $\phi_0$ to each candidate external graph, generating node embeddings and spherical coordinates. These serve as crucial metrics for identifying suitable external graphs to enhance anomaly detection tasks.

\vpara{Selection Criteria.}
To effectively select external data that enhances model training and enriches the embedding space, our selection mechanism adheres to two key criteria:
(1) \emph{Representativity}: 
As illustrated by the yellow triangles in Figure~\ref{fig: framework}(b), these data points are distributed near the blue points, demonstrating a high degree of similarity to the target graph.
(2) \emph{Diversity}: 
As depicted by gray triangles in Figure~\ref{fig: framework}(b), these data samples are widely scattered across the space, capturing diverse aspects of the embedding structure.
These two criteria are complementary: representativity ensures the relevance of the external data, while diversity maximizes the breadth of normal patterns represented. The ideal external data satisfies both, as exemplified by the half-yellow, half-gray triangle in Figure~\ref{fig: framework}(b). By integrating these principles, we calculate a final score for each candidate graph dataset to determine its suitability for improving anomaly detection tasks.

\vpara{{Representativity.}} 
The selected external data must closely resemble the target data to ensure effective integration. Otherwise, insufficient similarity may introduce noise, causing a distributional shift and potentially impairing model performance. We quantify representativity by evaluating it from two key perspectives.

\vpara{(1) Sphere-center Restriction.} The position of the embedding space, after combining external data with the target graph, should not diverge too much from its original configuration. Specifically, the center of the learned spherical space should not shift excessively. To measure this, we calculate the Euclidean distance between the centers of the target graph's embedding space and the mixed space (which includes both the target and external data). It can be formalized as:
\begin{equation}
    \mathrm{s}_{\text {sim }}^{\text {c }}=\| \textbf{c}_\text{in}-\textbf{c}_{\text {mix }} \|_2 ,
\end{equation}
where 
$\textbf{c}_\text{in }=\frac{1}{n_\text{in}} \sum_{z_i \in \boldsymbol{Z}_{in} }z_i,$ 
$ \textbf{c}_{\text {mix }} ={\frac{1}{{n_\text{in}}+{n_\text{out}}} } \sum_{z_i \in \boldsymbol{Z}_{in}\cup \boldsymbol{Z}_{out} } z_i$,  
$\boldsymbol{Z}_{in}$ and  $\boldsymbol{Z}_{out}$ represent the node embeddings of the target graph and the external data, while $n_\text{in}$ and $n_\text{out}$ denote the number of nodes in the target graph and the external data.
A small value for $\mathrm{s}_{\text{sim}}^{\text{c}}$ indicates a minimal shift, meaning the external data has been successfully integrated without distorting the embedding space. Conversely, a larger value suggests a significant shift, which may adversely affect the model's performance in detecting anomalies.

\vpara{(2) Distributional Similarity.} 
To further assess representativity, we measure the similarity between the distribution of the target graph’s node embeddings and the distribution of the mixed node embeddings. 
Specifically, we use the Wasserstein Distance to quantify the discrepancy between the spherical coordinates of the target graph's embeddings $\boldsymbol{\rho}_\text{in}$, and that of the mixed embeddings $\boldsymbol{\rho}_\text{mix}$. The Wasserstein Distance represents the minimum work required to transform one distribution into the other, considering both the values and directions of the embeddings.  
Mathematically,
it is computed as follows:
\begin{equation}
    {s}_{\text {sim }}^{ \text{d }}={Warssenstein}\left(\mathbb P\left(\boldsymbol{\rho}_{\text {in }}\right) ,\mathbb P\left(\boldsymbol{\rho}_{\text {mix}}\right)\right) 
\end{equation}
where $\mathbb{P}(\cdot)$ represents the probability distribution of the embeddings. A smaller Wasserstein distance indicates that the mixed data's distribution is similar to the target graph’s distribution, ensuring that the external data aligns well with the original data.

\vpara{ Diversity.} While the representativity criterion ensures that the selected external data is similar to the target graph, it alone may not be sufficient. Relying solely on representativity can be misleading, especially if the chosen data is overfitted to specific patterns. In such cases, the model may become overly specialized in recognizing certain normal patterns, leading to a loss of generalizability. To address this issue, we introduce {diversity} to ensure that normal patterns are distributed more evenly across the embedding space. By promoting a more uniform spread of data samples, diversity enhances the model's generalizability, enabling it to capture a broader range of patterns and better perform on diverse data.

In spherical space, diversity is achieved when embedding points are distributed more uniformly across the hypersphere, rather than being concentrated in a small region. This uniform scattering ensures that the node samples cover a larger portion of the hypersphere, capturing a wider variety of patterns and characteristics.
To quantify this, we use the Minimum Hyperspherical Energy (MHE) as a quantitative measure. 
This approach is motivated by the well-known Thomson problem in physics~\cite{Thomson}, which seeks to distribute $N$ electrons on the surface of a unit sphere in such a way that their potential energy is minimized. The core of solving this problem lies in finding a uniform distribution that maximizes the equal spacing between particles, thereby achieving the lowest energy state of the system. In our context, a higher energy indicates a higher degree of redundancy (\emph{i.e.}, more clustering of the points), while a lower energy signifies a more uniform scattering of the points on the hypersphere,  which is essential for ensuring diversity.
Specifically, we define the hyperspherical energy in the space $S^d$ as follows:
\begin{equation}
    s_{div} = \int_{r_\text{min}}^{r_\text{max}} \underset {u,v \in \mathcal{M}\left(\mathcal{S}(r)\right)}  {\int_u \int_v} e(u, v)  \mathrm{d} v {d} u\mathrm{d}r .\\ 
\end{equation}
where $\mathcal{M}\left(\mathcal{S}(r)\right)$ is the set of probability measures on the spherical surface at radius $r$, and $u,v$ are the samples within this set.
$e(u,v)$ is the energy function that represents the potential energy between two nodes $u$ and $v$ on the hypersphere. This function should be a decreasing, distance-based real-valued function~\cite{gao2024distribution,liu2018learning,lin2020regularizing}, assigning higher energy to closer points and lower energy to more distant points. Consequently, the sum of $e(u,v)$  over all pairs of nodes in the graph embedding space represents the total energy of the dataset. 
Here we adopt a Gaussian potential kernel following~\cite{bachman2019learning,cohn2007universally} to quantify $e(u,v)$, defined as:
\begin{equation}
e(u, v) \triangleq e^{-t\|u-v\|_2^2}=e^{2 t \cdot u^{\top} v-2 t}, \quad t>0
\end{equation}
where $t$ is a hyper-parameter for the Gaussian potential kernel. 
Moreover, we define
$r_{\min }=\min \left(\left\|z_i-c\right\|, z_i \in Z_{\text {in }} \cup Z_{\text {out }}\right)$, 
$r_{\max }=\max \left(\left\|z_i-c\right\|, z_i \in Z_{\text {in }} \cup Z_{\text {out }}\right)$.  
 The integral over $r$ ensures that we account for diversity at multiple scales of the hypersphere, rather than limiting the calculation to a fixed radius (such as 
$r=1$). This multi-scale approach provides a more comprehensive measure of data diversity.

\vpara{Final Score Calculation.}
The proposed representativity and diversity criteria together serve as a powerful indicator for evaluating the quality of the external data. We calculate the final score for each candidate dataset, transforming the selection problem into an optimization problem as follows:
\begin{equation}\label{equ:j}
\text{minimize} \;\;  \mathcal J(G) = \eta \text{MEAN}(
      {\hat{s}_{sim}^\text{c} +\hat{s}_{sim}^\text{d} }) +(1-\eta) 
      \hat{s}_{div},
\end{equation}
where $ \eta \in [0,1]$ is a trade-off parameter and the small hat on the terms $\hat{s}_{sim}^c$, $\hat{s}_{sim}^d$ and $\hat{s}_{div}$ indicates that all the values are $z$-normalized. This ensures the objective is invariant to the original scales of the values. Given the allocated budget of $k$ graphs, we select the top-k external graphs $\mathcal{G}_o$ with the smallest scores as determined by $Eq. ~\ref{equ:j}$. 
\subsection{Training Strategy}\label{subsec:training}

The model is initialized with the parameters $\theta^*$ of $ \phi_0 $ that were trained on the target graph $ G_t $, and then we continue training the model using the selected external graphs $ \mathcal{G}_o $. However, directly applying external data poses a challenge, as it may introduce irrelevant information and cause the model forgetting the previous knowledge learned from $ G_t $. To mitigate this issue, we incorporate an additional regularization term into the training objective. It ensures the new parameters learned from the external graph remain close to the parameters previously obtained from the target graph. Therefore, the loss function for the training model is :
\begin{equation}
\mathcal{L} = \mathcal{L}_\text{ano} + \sum_i \frac{\tau}{2} F_{ii}\left(\theta_i-\theta_i ^{*}\right)^2,
\end{equation}
where $\mathcal{L}_\text{ano}$ is the anomaly detection task loss on the external data, the parameter $\tau$ controls the balance between leveraging external graph data and preserving prior knowledge from target data, $F$ is the fisher information matrix, it quantifies the importance of each model parameter with respect to the knowledge learned from $G_t$.
       
\begin{table*}[t]
\caption{statistic of raw external data in the wild, where * denotes average statics of 16 augmented versions.}
\label{tab:external_data}
\centering
\resizebox{0.88\textwidth}{!}{
 \footnotesize
    \setlength\tabcolsep{2pt} 
    \begin{threeparttable}
    \begin{tabular}{cccccp{8cm}}
     \toprule
        & \multicolumn{1}{c}{Domain}  
        & \multicolumn{1}{c}{Name } 
        & \multicolumn{1}{c}{num\_nodes} 
        & \multicolumn{1}{c}{num\_edges} \\
     \midrule
     \multirow{26}*{\setlength\tabcolsep{1pt}\rotatebox{90}{\textbf{external data}}} 
     & \multirow{14}*{\textcolor[rgb]{0.404,0.173,0.075}{\textbf{Co-purchase}}} 
     & P-Learning & 651762 & 24599888 & purchase transactions between customers and products for learning \\
     &  & P-Learning-augmented & 543135* & 14760788*
& same as above(P-Learning) \\

     &  & P-Electronic & 243091 & 8767392 & purchase transactions between customers and products for electronic \\
     &  & P-Electronic-augmented & 202576* & 5260526* &  same as above(P-Electronic) \\
     &  & P-Entertainment & 606291 & 31779304 & purchase transactions between customers and products for entertainment \\
     &  & P-Entertainment-augmented & 505242* & 19048777*
 &  same as above(P-Entertainment) \\
     &  & P-Household & 374523 & 20448012 & purchase transactions between customers and products for household \\
     &  & P-Household-augmented & 312102* & 12279150*
 &  same as above(P-Household) \\
     &  & P-Fashion & 297098 & 14351216	& purchase transactions between customers and products for fashion \\
     &  & P-Fashion-augmented & 247581* & 8592927*
	&  same as above(P-Fashion) \\
     &  & YelpNYC & 138662 & 336489504 & interaction relationship between users and businesses such as reviews \\
     &  & YelpNYC-augmented & 135804* & 306434502*
 &  same as above(YelpNYC) \\
     &  & YelpRes & 28391 & 35786268 & interaction relationship between users and restaurants such as reviews and ratings  \\
     &  & YelpRes-augmented & 25391* & 33186543* &  same as above(YelpRes)  \\    
      \cmidrule(lr){2-6}
     & \multirow{4}{*}{\textcolor[rgb]{0.765,0.4078,0.145}{\textbf{Social}}}
     & Reddit & 33434 & 198448 & a social network representing reply relationships
\\
    &  & Reddit-augmented & 28290* & 125242*
 &  same as above(Reddit)
    \\
     &  & Instgram & 11339 & 144010 & a social network representing
following relationships \\
    &  & Instgram-augmented & 10258* & 133498*
 &  same as above(Instgram) \\ 
     
     \cmidrule(lr){2-6}
     & \multirow{6}{*}{\textcolor[rgb]{0.878,0.690,0.169}{\textbf{Citation}}}  
     & Cora & 2708 & 10858 & citations between Machine Learning papers \\
     &   & Cora-augmented & 2291* & 6849*
 &  same as above(Cora) \\
     &   & Pubmed & 19717 & 88670 & citations between scientific papers \\
     &   & Pubmed-augmented & 16683* & 55742*
 &  same as above(Pubmed) \\
     &   & Arxiv	 & 169343 & 1166243 & citations between papers on the arxiv \\
     &   & Arxiv-augmented & 143290* & 736371*
 &  same as above(Arxiv) \\ 
     \cmidrule(lr){2-6}
     & \multirow{2}{*}{\textcolor[rgb]{0.9,0.2,0.3294}{\textbf{Hyper link}}} 
     & Wikics & 11701 & 431726 & citations between Wikipedia articles \\
     & & Wikics-augmented & 9901* & 271760*
 &  same as above(Wikics) \\

\bottomrule
    \end{tabular}
    \end{threeparttable}
 }
\end{table*}
\begin{table*}[t]
\caption{statistics of target anomaly data}
\label{tab:target_data}
\centering
\resizebox{0.85\textwidth}{!}{
 \footnotesize
    \setlength\tabcolsep{2pt} 
    \begin{threeparttable}
    \begin{tabular}{cccccccc}
     \toprule
        & \multicolumn{1}{c}{Domain}  
        & \multicolumn{1}{c}{Name } 
        & \multicolumn{1}{c}{num\_nodes} 
        & \multicolumn{1}{c}{num\_edges}
        & \multicolumn{1}{c}{num\_normals}  
        & \multicolumn{1}{c}{num\_abnormals}  
        & \multicolumn{1}{c}{abnormal\_ratio}  \\
     \midrule
     \multirow{6}*{\rotatebox{90}{\textbf{target data}}} & 
     \multirow{2}{*}{\textcolor[rgb]{0.5,0.5,0.2}{\textbf{co-purchase}}} 
     & YelpHotel & 4472 & 709480 & 4264 & 208 & 4.65$\%$ \\
     &  & AmazonCN & 7778 & 6804032 & 7389 & 389 & 5.00$\%$ \\
     
     & \multirow{2}{*}{\textcolor[rgb]{0.404,0.173,0.075}{\textbf{Social-bot}}} & C-15 & 1890 & 8494 & 1219 & 671 & 35.50$\%$ \\
     &  & Twitter-20 & 11345 & 12816 & 4269 & 6393 & 56.35$\%$ \\
     \cmidrule(lr){2-8}
     & {\textcolor[rgb]{0.765,0.4078,0.145}{\textbf{Work Collaboration}}} 
     & Tolokers & 11758 & 1038000 & 9192 & 2566 & 21.82$\%$ \\
     \cmidrule(lr){2-8}
     & {\textcolor[rgb]{0.878,0.690,0.169}{\textbf{Email-Communication }}}  
    & Enron & 13533 & 176987 & 13528 & 5 & 0.04$\%$ \\

\bottomrule
    \end{tabular}
    \end{threeparttable}
 }
\end{table*} 
\subsection{Extension with labeled candidate data}
In some practical scenarios where there are external anomaly detection datasets and abnormal labels the external data is available, our framework can also be adopted. Such external signal can help to adopt to real-world anomalous to some degree. Specifically, when conducting data selection, we measure the representativity and diversity for labeled normal samples and abnormal samples respectively, taking their average as the final score.

\section{EXPERIMENT} \label{sec:exp}
In this section, we mainly evaluate the effectiveness of Wild-GAD to answer the following two questions:
(1) Does the selection strategy employed in our framework effectively measure the quality of the external data? (2) Does the performance of the anomaly detection task on the target graph improve with the selected external data? (3) How much labeling effort can be saved by using our framework?

\subsection{Experimental Setup}
\textbf{Datasets.} 
We collect six real-world graph datasets with organic anomalies from a wide spectrum of domains (See the bottom part of Table~\ref{tab:target_data} for detailed statistics), namely C15~\cite{cresci2015fame}, AmazonCN~\cite{wu2023neighbors}, YelpHotel~\cite{mukherjee2013yelp}, Twitter-20~\cite{feng2021twibot}, Tolokers~\cite{platonov2023critical} and Enron~\cite{klimt2004introducing}. C-15 is a dataset composed of genuine and fake Twitter accounts, which are manually annotated. AmazonCN is a proprietary data collected from Amazon China. It consist of products information, users and corresponding Chinese review. YelpHotel features real hotel reviews from Yelp.com, curated through Yelp's proprietary filtering mechanism. Twitter-20 is a comprehensive sample of the Twittersphere and it is representative of the current generation of Twitter bots and genuine users. Tolokers is based on data from the Toloka crowdsourcing platform. The nodes represent tolokers (workers). Enron is a dataset of emails generated by employees of the Enron Corporation. More details can be seen in Appendix A.

\begin{table*}
\caption{Comparison of AUC-PR of different graph anomaly detection methods; OOM means out of memory, ``OC'' means one-class backbone, ``GAE'' means auto-encoder backbone, ``s'' and ``M'' means the external data budget is 1 and 2 respectively, ``UL'' and ``L'' means whether external data are labeled.}
 \centering
 \resizebox{1.9\columnwidth}{!}{\begin{tabular}{cc|cccccc}
\hline
                       & { Method}           & { YelpHotel}                  & { AmazonCN}                   & { C-15}                       & { Twitter-20}                 & { Tolokers}                   & { Enron}                      \\ \hline

{ }                           & { { ANOMALOUS}}  & { 5.99(0.77)}                 & { 10.21(8.34)}                & { 31.01(1.06)}                   & { 53.76(0.10)}                 & { 15.32(0.09)}                & { 0.07(0.04)}                 \\
{ }                           & { { Radar}}      & { 5.18(0.00)}                 & { 10.80(0.00)}                 & { 31.15(0.00)}                & { 53.53(0.00)}                & { 15.4(0.00)}                 & { 0.06(0.01)}                 \\
{ }                           & { { DOMINANT}}   & { 3.98(0.10)}                  & { 6.34(4.38)}                 & { 54.61(0.21)}                & { 53.85(0.86)}                & { 25.51(0.29)}                & { 0.06(0.02)}                    \\
{ }                           & { { AnomalyDAE}} & { { 4.66(0.82)}}           & { 11.09(2.48)}                & { 36.2(2.87)}                & { 57.96(3.63)}                & { 20.98(4.23)}                & { 0.05(0.01)}                 \\
{ }                           & { { AdONE}}      & { { 4.43(0.08)}}           & { 3.11(0.04)}                 & { 33.82(2.31)}                 & { 57.13(1.63)}                & { 29.03(2.00)}                   & { 0.07(0.01)}                 \\
{ }                           & { { GAAN}}       & { { 3.68(0.18)}}           & { 2.88(0.12)}                 & { 27.57(0.61)}                & { 60.05(0.47)}                & { 27.77(1.89)}                & { {0.12(0.15)}}        \\
{ }                           & { { CoLA}}       & { { 4.77(0.15)}}           & { 5.75(1.40)}                  & { 45.26(2.95)}               & { 58.96(5.57)}                & { 21.98(1.25)}                & { 0.04(0.01)}                 \\
{ }                           & { { CONAD}}      & { { 4.07(0.14)}}           & { OOM}                        & { 53.76(1.93)}                & { 53.77(0.75)}                & { 25.58(0.42)}                & { 0.06(0.01)}                    \\
{ }                           & { { DONE}}       & { { 4.17(0.18)}}           & { 10.82(6.67)}                & { 29.29(2.86)}                & { 59.74(1.09)}                & { 26.83(1.28)}                & { {0.12(0.15)}}        \\
{ }                           & { { GAE}}        & { { 6.09(0.69)}}           & { 5.04(0.80)}                  & { 21.67(0.57)}                & { 58.84(1.35)}                & { 24.87(1.08)}                & { 0.05(0.01)}                 \\
{ }                           & { { OCGNN}}      & { 7.04(0.31)}                 & { 5.36(2.74)}                 & { 27.41(1.01)}                & { 63.52(1.46)}                & { 19.57(3.95)}                & { 0.09(0.03)}          \\ & TAM & { 7.07(0.03)} & 	3.16(0.02)	& 66.27(0.39)& 	52.01(0.00)& 	26.68(0.03)& 	0.04(0.00)     \\ 
& RAND & 3.78 (0.21) &	3.43(0.00)	 & 51.79(0.01) &	56.35(0.10)	 & 28.96(0.35) &	0.05(0.00)\\
\cline{2-8} 
\multirow{-12}{*}{\rotatebox[origin=c]{90}{AUC-PR}}  & 
{ { Wild-GAD(OC/S/UL)}}       & { { {8.65(1.02)
}}}   & { {31.99(1.24)
}}       & { {86.85(3.32)}}       & { {62.32(1.12)

}}       & { {28.93(1.35)

}}       & { {0.13(0.02)
}}        
\\ & { { Wild-GAD(OC/M/UL)}}       & { { \textbf{8.84(0.64)}}}   & { \textbf{35.38(0.54)}}       & { \textbf{89.24(2.45)
}}       & { {64.23(1.25)}}       & { {30.21(1.21)}}       & { \textbf{0.14(0.13)}}  
\\ & { { Wild-GAD(GAE/S/UL)}}    & { { {8.21(0.12)}}}   & { {33.38(0.63)}}       & { {85.72(1.78)}}       & { {70.62(2.25)}}       & { {30.79(0.89)}}       & { {0.12(0.05)}}     
\\& { { Wild-GAD(GAE/M/UL)}}       & { { {8.35(1.53)}}}   & { {34.28(1.24)}}       & { {87.72(1.28)}}       & { \textbf{72.42(1.05)}}       & { \textbf{32.79(1.39)}}       & { {0.13(0.05)}} 
\\  \hline
& { { Wild-GAD(OC/S/L)}}       & { { {9.32(0.43)}}}   & { {35.65(1.23)}}       & { {85.45(0.23)}}       & { {65.12(1.02)}}       & { {31.61(1.23)}}       & { {0.14(0.08)}} 
\\ & { { Wild-GAD(OC/M/L)}}       & { { \textbf{9.75(0.34)}}}   & { \textbf{38.64(1.20)}}       & { {87.82(1.43)}}       & { {67.23(0.89)}}       & { {32.52(2.83)}}       & { {0.13(0.05)}} 
\\ & { {Wild-GAD(GAE/S/L)}}       & { { {9.14(2.35)}}}   & { {37.64(0.57)}}       & { {87.45(1.21)}}       & { {73.68(1.37)}}       & { {33.21(1.80)}}       & { \textbf{0.15(0.03)}}
\\ & { { Wild-GAD(GAE/M/L)}}       & { { {9.31(1.27)}}}   & { {38.12(0.67)}}       & { \textbf{88.82(1.04)}}       & { \textbf{74.13(2.24)}}       & { \textbf{34.52(0.46)}}       & { {0.14(0.04)}}
\\ \hline
\label{tab:ap}
\end{tabular}}
\end{table*}
\vpara{Baselines.}
We compare our proposed method with nine unsupervised graph anomaly detection models, including DOMINANT~\cite{DOMINANT}, AnomalyDAE~\cite{AnomalyDAE}, AdONE~\cite{AdONE}, GAAN~\cite{GAAN}, CoLA~\cite{CoLA}, CONAD~\cite{CONAD}, DONE~\cite{AdONE}, GAE~\cite{GAE}, OCGNN~\cite{OCGNN}, TAM~\cite{TAM} and RAND~\cite{RAND}.
\begin{table*}
\vspace{-0.1 in}
\caption{Comparison of AUC-ROC of different graph anomaly detection methods; OOM means out of memory, ``OC'' means one-class backbone, ``GAE'' means auto-encoder backbone, ``s'' and ``M'' means the external data budget is 1 and 2 respectively, ``UL'' and ``L'' means whether external data are labeled.}
 \centering
 \resizebox{1.9\columnwidth}{!}{\begin{tabular}{cc|cccccc}
\hline
                       & { Method}           & { YelpHotel}                  & { AmazonCN}                   & { C-15}                       & { Twitter-20}                 & { Tolokers}                   & { Enron}                      \\ \hline
                       & { ANOMALOUS}        & { 48.88(2.35)}                & { 54.99(9.63)}               & { 16.08(0.89)}                & { 41.52(0.05)}                & { 28.47(0.5)}                 & { 50.00(8.76)}                \\
                        & { { Radar}}      & { { 46.74(0.00)}}          & { { 56.65(0.00)}}          & { { 15.61(0.00)}}          & { { 42.14(0.00)}}          & { { 28.8(0.00)}}           & { { 52.54(0.00)}}          \\
                       & { { DOMINANT}}   & { 44.81(1.17)}                & { 42.30(8.25)}               & { 67.49(0.51)}                & { 41.62(2.57)}                & { 48.05(0.42)}                & { 47.85(3.72)}                \\
                      & { { AnomalyDAE}} & { { 47.53(4.13)}}          & { { 76.93(4.94)}}          & { { 53.98(3.31)}}         & { { 48.98(5.16)}}          & { { 46.91(7.66)}}          & { { 53.17(2.53)}}          \\
                       & { { AdONE}}      & { { 46.81(0.59)}}          & { { 23.82(0.98)}}          & { { 52.74(4.76)}}          & { { 45.28(3.20)}}           & { { 53.96(3.10)}}           & { { 43.87(4.51)}}          \\
                    & { { GAAN}}       & { { 40.85(1.68)}}          & { { 19.61(3.61)}}          & { { 33.53(1.21)}}          & { { 50.44(0.45)}}          & { { 56.11(1.74)}}          & { { 58.19(8.72)}}          \\
                         & { { CoLA}}       & { { 49.83(1.48)}}          & { { 54.38(11.31)}}         & { { 58.04(2.35)}}         & { { 50.02(8.14)}}          & { { 49.33(1.80)}}           & { { 42.03(16.65)}}         \\
{ }                           & { { CONAD}}      & { { 45.94(1.59)}}          & { { OOM}}                  & { { 78.61(2.27)}}          & { { 41.32(2.71)}}          & { { 48.18(0.63)}}          & { { 47.1(0.80)}}            \\
{ }                           & { { DONE}}       & { { 45.71(1.43)}}          & { { 61.9(13.19)}}          & { { 42.05(7.18)}}          & { { 50.03(1.60)}}           & { { 52.27(1.38)}}          & { { 45.69(2.51)}}          \\
{ }                           & { { GAE}}        & { { 58.29(3.06)}}          & { { 49.27(9.17)}}          & { { 11.02(5.65)}}          & { { 50.3(3.26)}}           & { { 49.27(1.75)}}          & { { 41.91(10.35)}}         \\
{ }                           & { { OCGNN}}      & { 59.44(1.20)}                 & { 45.42(12.3)}                & { 36.90(3.85)}                 & { 53.70(1.41)}                & { 43.83(6.80)}                 & { 43.42(9.10)}                 \\& TAM & 62.51(0.17)&	22.57(0.33)&	83.48(0.34)&	36.43(0.01)	&60.35(0.13)	& 45.24(0.01)     \\ 
& RAND & 42.59(0.03)&	33.51(0.01)&	75.92(0.01)	&42.58(0.27)&	56.85(0.59)&	39.95(3.09)\\ \cline{2-8} 
\multirow{-12}{*}{\rotatebox[origin=c]{90}{AUC-ROC}}  & 
{ { Wild-GAD(OC/S/UL)}}       & { { {{66.11(1.20)}}}}   & { {80.03(2.57)}}       & { {97.22(1.23)}}       & { {54.72(2.32)
}}       & { {54.35(3.80)
}}       & { \textbf{66.08(3.10)
}}        
\\ 
& { { Wild-GAD(OC/M/UL)}}       & { { \textbf{68.55(1.22)}}}   & { \textbf{81.43(1.23)}}       & { {\textbf{97.72}(1.78)}}       & { {57.68(1.35)}}       & { {60.19(0.45)}}       & { {64.12(0.25)}}  
\\ & { { Wild-GAD(GAE/S/UL)}}       & { { {62.31(0.28)}}} & { { {76.04(1.23)}}} & { { {95.82(1.23)}}} & { { {63.29(1.24)}}} & { { {65.52(1.83)}}} & { { {63.65(2.05)}}}   
\\& { { Wild-GAD(GAE/M/UL)}}       & { { {64.20(0.13)}}}   & {
{77.38(0.45)}}       & { {96.62(1.92)}}       & 
{ \textbf{65.62(1.35)}}       & { \textbf{67.29(0.16)}}       & { {65.65(1.23)}} 
\\ \hline
& {{Wild-GAD(OC/S/L)}}  & {{{68.32(1.32)}}} &{{{82.14(2.31)}}} &{{{97.45(1.23)}}} &{{{58.32(1.32)}}} &{{{55.61(1.23)}}} &{{\textbf{64.42(1.32)}}} 
\\ & {{Wild-GAD(OC/M/L)}}  & {\textbf{71.31(0.28)}} &{\textbf{84.64(1.89)}} &{\textbf{98.82(1.13)}} &{{{64.13(1.24)}}} &{{{60.52(1.83)}}} &{{{64.02(2.05)}}}
\\ & {{Wild-GAD(GAE/S/L)}}  & {{{64.94(1.35)}}} &{{{78.64(2.57)}}} &{{{96.45(1.03)}}} &{{{67.68(2.32)}}} &{{{66.21(2.80)}}} &{{{63.42(1.18)}}} 
\\ & {{Wild-GAD(GAE/M/L)}}  & {{{66.31(2.34)}}} &{{{79.04(1.23)}}} &{{{97.82(1.04)}}} &{{\textbf{70.13(1.24)}}}  &{{\textbf{67.42(0.93)}}} &{{{65.02(2.12)}}} 
\\ \hline
\label{tab:auc}
\end{tabular}}
\end{table*}

\vpara{Evaluation Metrics.}
In order to evaluate the performance of our proposed method and baselines, we adopt two popular and complementary evaluation metrics for anomaly detection,
Area Under the Receiver Operating Characteristic Curve (AUROC)  and Area Under the precision recall curve (AUCPR). Higher AUROC/AUCPR indicates better performance. All of our results are reported as the averaged performance with standard deviation of  five runs.

\vpara{Implementation details.} 
When constructing our database, UniWildGraph (introduced in Section~\ref{subsec:dataset}), we align the feature space using mBERT with a maximum token length of 512.
For data selection, we set the trade-off parameter $\eta$ to 0.5. For the Wasserstein Distance calculation, we follow the default hyperparameters specified in ~\cite{WassersteinDistance}.
For one-class model training, we set the parameter $\beta$ to 0.5. For the baseline methods, we initialize them with the same parameters suggested by their default settings.
{Apart from one-class, we also adapt our framework to the auto-encoder model and the parameters are set as the same as the default in~\cite{GAE}.} More technique analysis about this backbone model can be found in Appendix~\ref{app:GAE}.

All baseline methods are implemented using the PyGOD package~\cite{JMLR:v25:23-0963}, which is specifically designed for graph anomaly (outlier) detection. The learning rate for each method (except OCGNN) is set to 0.1, and the number of epochs is set to 100. For OCGNN, we set the learning rate to 0.001. The other parameters remain the same as the default settings in PyGOD.
To ensure successful execution of experiments on the C15 dataset, in addition to the default hyperparameters, we set the hidden dimension to 128. Due to the complexity of the AmazonCN dataset, we set the number of neighbors in sampling to 5, alongside the base hyperparameter settings.   Our codes are available at https://github.com/zjunet/
Wild-GAD.

\subsection{Experimental results}
\vpara{Effectiveness of Wild-GAD. }
We evaluate our framework based on two backbone models (\emph{i.e.}, One-Class (OC) based Model and Graph-AutoEncoder (GAE)based Model)  denoted as Wild-GAD(OC) and Wild-GAD(AE) respectively. 
We conduct experiments with the selection budget  (\emph{i.e.}, the number of datasets selected) of 1 ( denoted as Wild-GAD(S))and 2( denoted as Wild-GAD(M))  
What's more, We also test the performance of our framework with exposure to external labeled anomaly detection datasets. We use  Wild-GAD(UL) to represent the setting that all the external data are unlabeled and Wild-GAD(L) to represent the external label setting.

Table~\ref{tab:auc} and Table~\ref{tab:ap} show the overall AUCROC and AUCPR  results on six real-world GAD datasets.  
From the results, We have the following observations that 
(1) our framework Wild-GAD outperforms the baseline methods, having an average {18\%} AUCROC and {32\%} AUCPR improvement over the best-competing methods. The key reason behind the performance improvement can be attributed to the proper involvement of external data. 
(2) We find that in most cases, adding more external data can lead to higher performance improvements in anomaly detection compared to adding just one data source, which means more diverse normal patterns have been seen during training.
(3) It can be found that in some cases, compared to only unlabeled external data, the performance of using labeled external data has little improvement sometimes and even drops (\emph{e.g.,} dataset Enron and dataset Tolokers). It can be explained by the abnormal behavior patterns of the external abnormal signal sometimes can be largely different from that of target data.  As our labeled external data Instagram is from social network, YelpNYC and YelpRes are co-purchase network, while Enron is an email network and Tolokers are workers network. 

\begin{figure}[h]
    \centering
    \includegraphics[width=1.0\columnwidth]{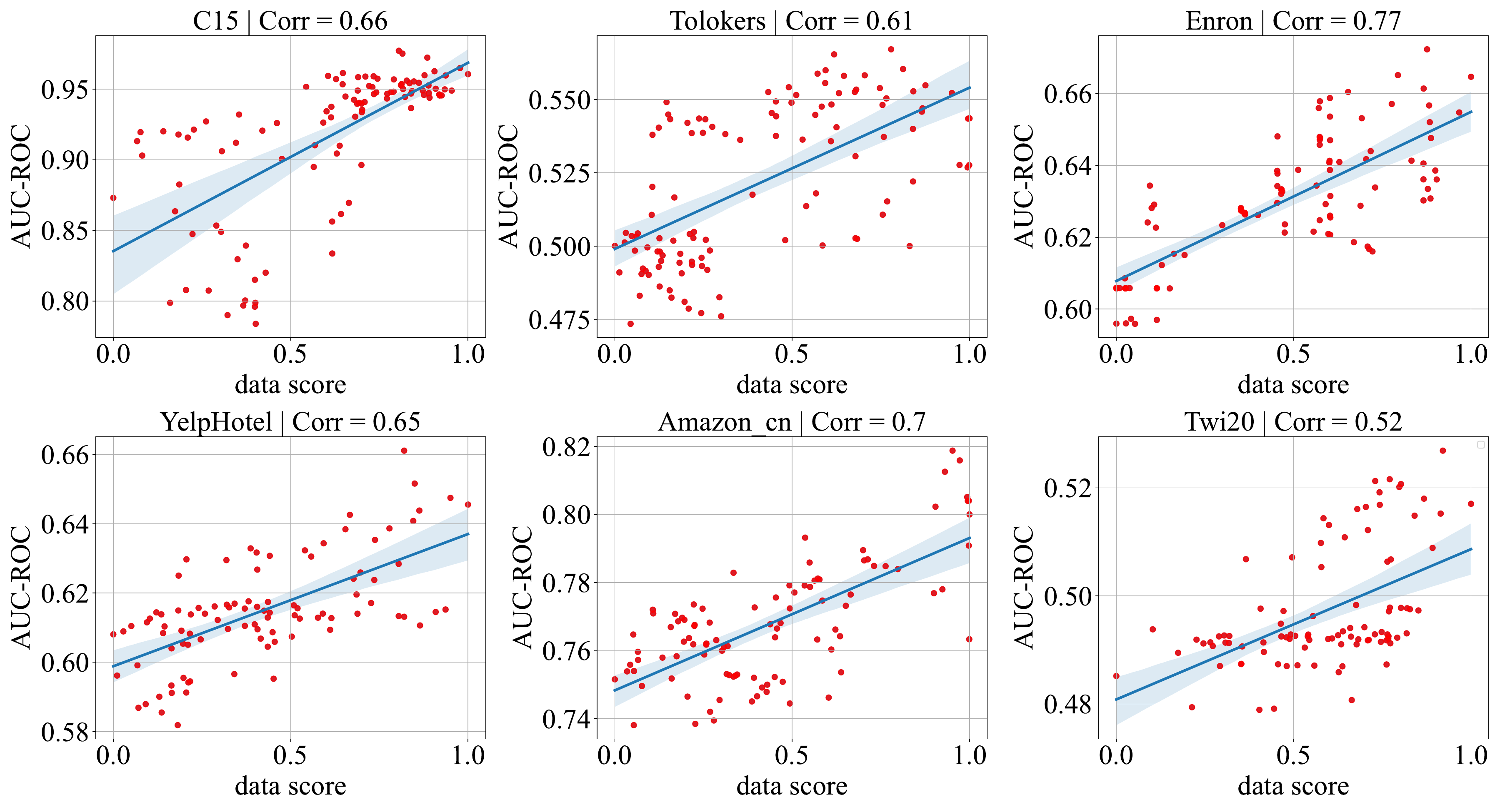}
         \vspace{-0.2in}
    \caption{The score of external data vs. the detection performance (AUC-ROC) on different datasets}
     \vspace{-0.2in}
    \label{fig:corre_occ}
\end{figure}
\begin{figure}[h]
    \centering
    \includegraphics[width=1.0\columnwidth]{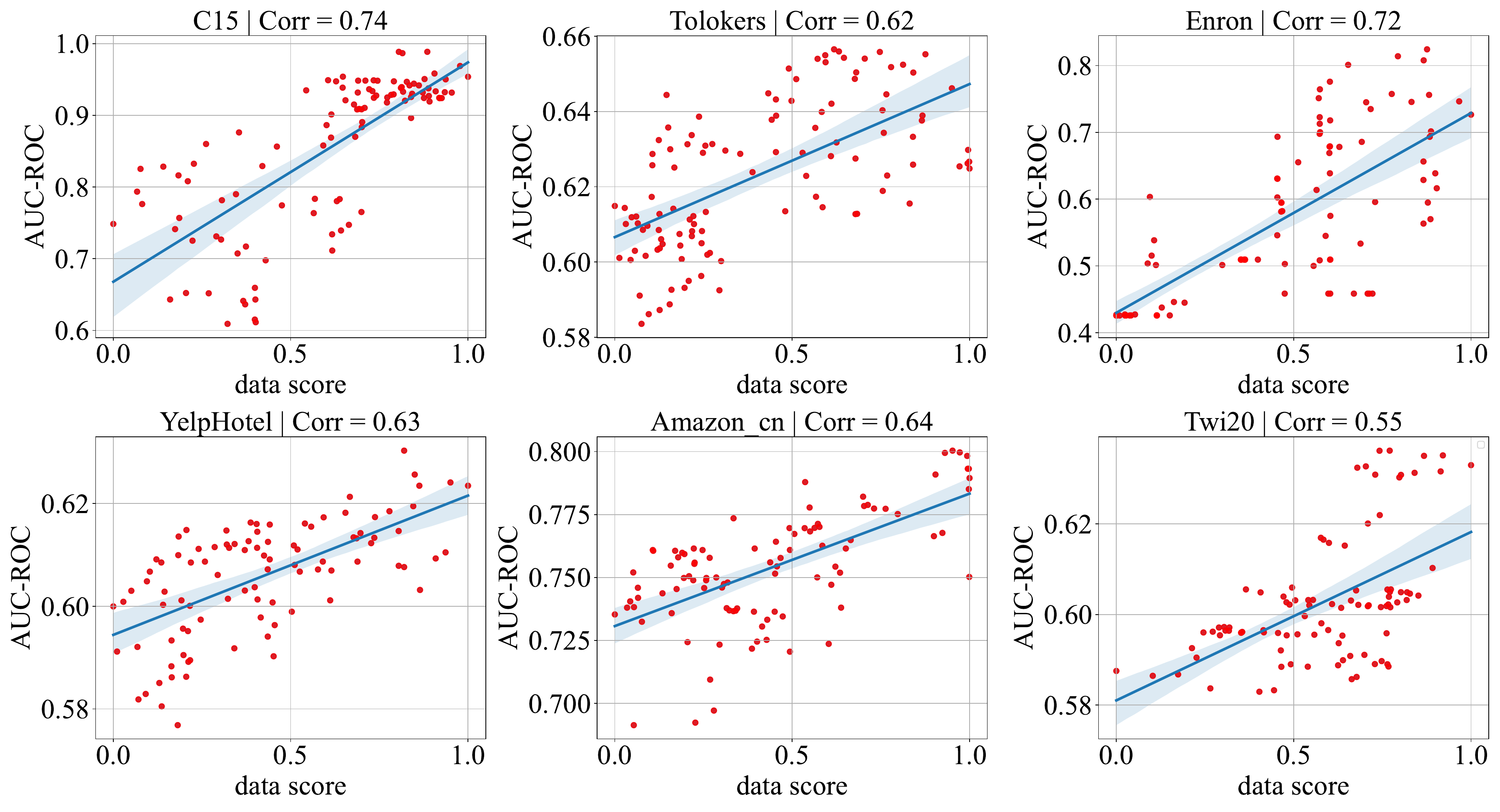}
    \vspace{-0.2in}
    \caption{The score of external data vs. the detection performance (AUC-ROC) on different datasets}
    \vspace{-0.1in}
    \label{fig:corre_gae}
\end{figure}
\vpara{Effectiveness of selection strategy.}
We evaluate Pearson correlation coefficient between
the AUROC performance and the external data score. A higher coefficient indicates a better estimation. Figure~\ref{fig:corre_gae} shows our estimated external data score
(in x-axis) versus the best AUC performance based on WilD-GAD(OC/S/UL) (in y-axis) of all <external data, target data>
pairs (one point represents the result of one pair). Another correlation display based on GAE backbone can be found in Appendix B Figure~\ref{fig:corre_gae}.
We
find that there is a strong positive correlation between estimated
data score and the performance on
all  datasets, which also suggests the significance of
our strategy of selection.

\begin{figure} [ht] 
    \centering
   \vspace{-0.1in}
    \includegraphics[width=0.9\columnwidth]{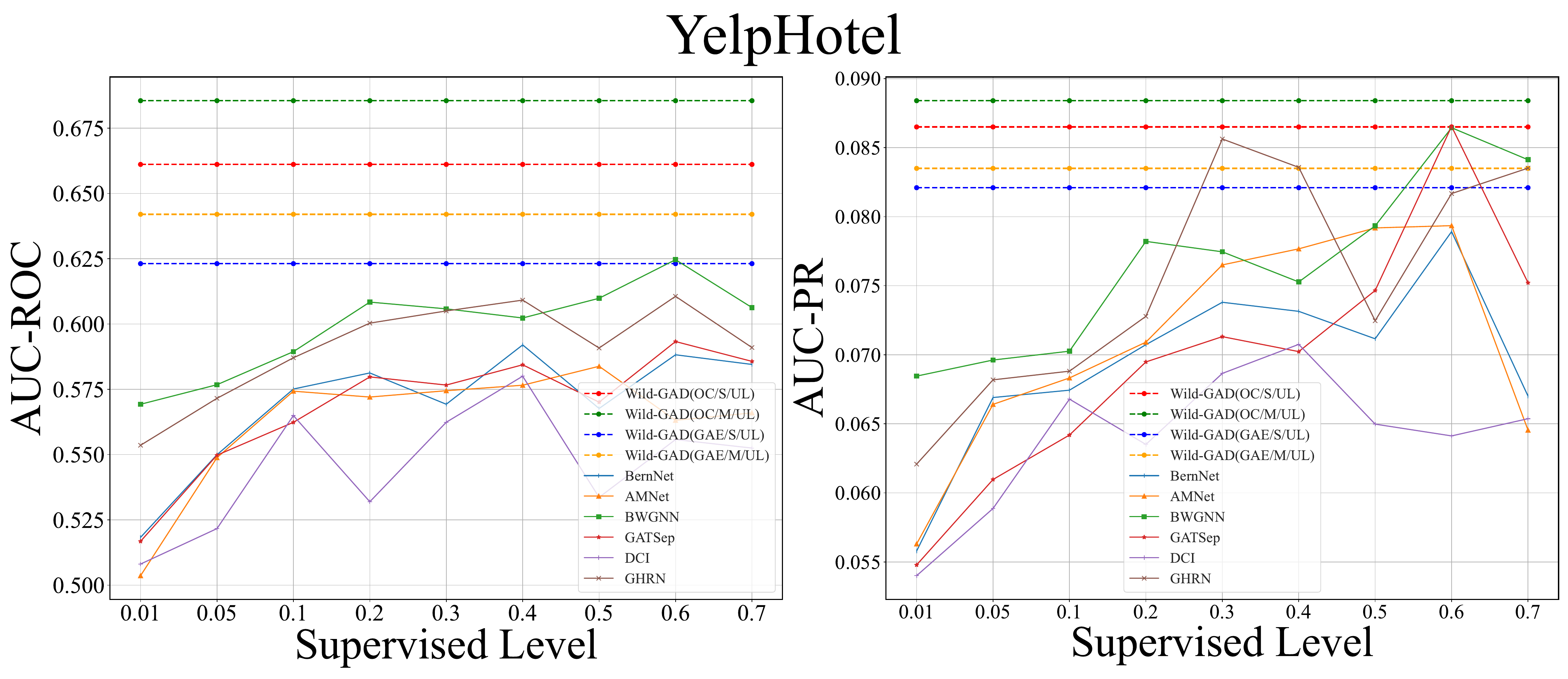}  
     \vspace{-0.1in}  
    \captionsetup{font={small}}
    \caption{Comparison with semi-supervised methods under varying supervision on YelpHotel.}  
    \label{fig:reg}
\end{figure}

\vpara{Performance Comparison with Semi-Supervised Methods.}
To assess the performance of our framework under different levels of supervision, we extended our comparison to encompass a range of label ratios: $1\%, 3\%, 5\%, 10\%, 40\%, 50\%,60\%  $ and $70\%$ of different semi-supervised methods 
BernNet~\cite{BernNet}, AMNet~\cite{AMNet}, BWGNN~\cite{BWGNN}, GATSep~\cite{GATSep}, DCI~\cite{DCI},  and GHRN~\cite{GHRN}.
The outcomes are presented in Figure~\ref{fig:reg}. We note that Wild-GAD outperforms its counterparts at label ratios up to 70\% based on both AUCROC and AUCPR metrics, demonstrating the effectiveness of leveraging external data for graph anomaly detection.

 \vspace{-0.2in}
\section{CONCLUSION}
In this study, we introduce Wild-GAD, a framework that leverages external graph data through an unsupervised paradigm to address the challenges in GAD. By developing the UniWildGraph database, which offers diverse domain coverage and a unified feature space, we enhance the capability of GAD models to learn the patterns of normality more effectively. The design of the strategic data selection process based on representativity and diversity ensures the relevance and utility of external data for improving anomaly detection performance on specific target graphs. Extensive experiments on multiple test datasets demonstrate the effectiveness of our proposed method.
\begin{acks}
This work was supported in part by NSFC (62206056, 92270121, 62176233, 62322606, 62441605) and CIPSC-SMP-Zhipu Large Model Cross-Disciplinary Fund (ZPCG20241030332).
Carl Yang was not supported by any grants from China.
\end{acks}
\newpage
\bibliographystyle{ACM-Reference-Format}
\balance
\bibliography{refer}

\appendix
\newpage
\appendix
\section {Datasets Details}\label{app:dataset}
In YelpHotel and AmazonCN, users are represented as nodes, and an undirected edge is established between them if two users have both commented on the same hotel or products. The embedding of each user is calculated as the average on vectors of every text-based review. It should be noted that the vectorization utilizes the same feature alignment approach introduced in Section~\ref{subsec:dataset} to unify the feature space. To simulate the true situation of business context, we control the proportion of abnormal nodes as around 5\%. Also, we take the maximal connected subgraph of the resulted graph for usage. In C-15 and Twitter-20, a directed edge is created between users when a user follows another user. We filter out users without comments and use the same method in co-purchase datasets to calculate user embedding. In Tolokers, an edge connects two tolokers if they have worked on the same task. In Enron, if an address sends at least one email to another address, the graph contains an undirected edge between them. We use the same feature alignment approach to get node embbedings and get maximal connected subgraphs for usage in these two datasets.

The Twitter-20 dataset used in our study is initially obtained from~\cite{feng2021twibot}, which is designed for Twitter bot detection. It contains 5,237 humans, 6,589 bots (anomalies). In order to preserve the dense graph structure follow relationship forms, the authors also supply 217,754 unsupervised users (with no labels). In our experiments, we classify bots as anomalies and apply Twitter-20 for anomaly detection in social networks. It is worth noting that unsupervised users are included in the training process, while evaluation is based solely on labeled samples. To maintain dataset integrity, we remove inactive users (those who have no posts, followers, or followings). We then organize the remaining users into a complete graph without further processing, preserving the original characteristics of Twitter-20. In the final version of Twitter-20 used for our study, there are 4,269 humans, 6,393 bots, and 683 unsupervised users. This results in a higher proportion of anomalies in Twitter-20 compared to normal users.
\section{Discussion about graph data in the wild }\label{app:graph-wild}
In general, "data in the wild" refers to real-world data collected from natural environments characterized by diversity, unstructured and inconsistent formats, and label scarcity. This data typically comes from various sources and domains, making it more challenging to process compared to clean, curated datasets~\cite{Gao2024LearningIT}. In our context, "graph data in the wild" refers to external graph data sourced from multiple real-world domains, including e-commerce, social networks, citation networks, and hyperlink networks. This aligns with findings in the literature that emphasize the need for diverse data sources to enhance model performance~\cite{BWGNN}. To enhance our collected graph data in the wild, we applied several augmentation techniques with varying ratios to increase diversity. More importantly, similar to the survey literature on data in the wild, graph data in the wild also exhibits unstructured and inconsistent formats, and label scarcity characteristics. 
Accordingly, to address the heterogeneity of features across these datasets, we performed unified feature processing. Since most graph data in the wild lacks labels, our framework focuses on leveraging unlabeled data, making unsupervised techniques crucial for effectively utilizing this data.

\section{Regarding the extra noise introduced by external data}\label{app:noise}
We have considered the potential extra noise introduced by external data and correspondingly adopt several strategies. 
We introduce a representativity criterion during the selection of external data, focusing on the similarity between the external graphs and the target graph. This criterion helps ensure that the selected external data shares common characteristics with the target data, thereby reducing the likelihood of incorporating irrelevant or misleading information. What's more, the application scope (which target dataset can potentially benefit from our work) of our framework can be characterized by the threshold of our calculated data score. Target data that does not meet this threshold is excluded from consideration, thereby avoiding potential biases introduced by irrelevant external data and preventing a decrease in model performance.
\begin{table*}[!ht]
\caption{Ablation Study AUC-PR Comparisons}
\vspace{-0.1in}
 \centering
 \resizebox{1.5\columnwidth}{!}{\begin{tabular}{lcccccc}
\toprule
{Method} & {YelpHotel} & {AmazonCN}  & {C-15}&{Twitter-20} & {Tolokers} & {Enron} \\ \hline
Wild-GAD-R(OC/S) & 4.54(1.34) & 18.72(1.98) & 68.66(1.54) & 61.12(1.06) & 25.17(2.34) & 0.10(0.02) \\ 
Wild-GAD-D(OC/S) & 7.34(1.62) & 24.52(1.87) & 84.23(2.31) & 59.13(0.98) & 25.12(1.23) & 0.11(0.02) \\ 
Wild-GAD(OC/S) & 8.65(1.02) & 31.99(1.24) & 86.85(3.32) & 62.32(1.12) & 28.93(1.35) & 0.13(0.02) \\ \hline
Wild-GAD-R(OC/M) & 5.86(2.60) & 17.13(2.25) & 67.62(2.08) & 54.24(1.56) & 23.65(2.43) & 0.09(0.02) \\ 
Wild-GAD-D(OC/M) & 6.76(0.31) & 23.12(0.43) & 83.12(1.54) & 53.12(3.12) & 24.12(1.14) & 0.12(0.03) \\ 
Wild-GAD(OC/M) & \textbf{8.84(0.64)} & \textbf{35.38(0.54)} & \textbf{89.24(2.45)} & \textbf{64.23(1.25)} & \textbf{30.21(1.21)} & \textbf{0.14(0.13) }\\ 
\bottomrule
\label{tab:ablation1}
\end{tabular}}
\end{table*}

\begin{table*}[!ht]
\caption{Ablation Study AUC-ROC Comparisons}
\vspace{-0.1in}
 \centering
 \resizebox{1.5\columnwidth}{!}{\begin{tabular}{lcccccc}
\toprule
{Method} & {YelpHotel} & {AmazonCN}  & {C-15}&{Twitter-20} & {Tolokers} & {Enron} \\ \hline
Wild-GAD-R(OC/S) & 52.13(2.12) & 70.34(2.38) & 89.13(2.32) & 55.09(2.45)&53.32(3.45) & 62.87(2.98) \\ 
Wild-GAD-D(OC/S)  & 61.72(1.87) & 76.48(2.51) & 93.62(1.83) & 53.98(1.31)&52.99(2.12) & 63.81(1.70) \\ 
Wild-GAD(OC/S)  & 66.11(1.20) & 80.03(2.57) & 97.22(1.23) & 54.72(2.32)&54.35(3.80) & \textbf{66.08(3.10)} \\ \hline
Wild-GAD-R(OC/M) & 53.46(1.95) & 71.91(2.45) & 87.24(3.09) & 55.15(2.67) &52.38(2.45) & 61.98(2.01) \\ 
Wild-GAD-D(OC/M)& 64.43(1.49) & 77.12(1.45) & 94.65(2.01) & 51.13(1.42)&55.12(2.56) & 63.03(0.98) \\ 
Wild-GAD(OC/M) & \textbf{68.55(1.22)} & \textbf{81.43(1.23) }& \textbf{97.72(1.78) }& \textbf{57.68(1.35)}&\textbf{60.19(0.45)} & 64.12(0.25) \\ 
\bottomrule
\label{tab:ablation2}
\end{tabular}}
\end{table*}
\section{Ablation Studies }\label{app:ablation}
To evaluate the effectiveness of our criteria, we conduct ablation studies on the absence of representativity and Diversity criteria, as Wild-GAD-R and Wild-GAD-D respectively based on one-class model. As shown in Table~\ref{tab:ablation1} and Table~\ref{tab:ablation2}, we found that performance declines in both cases without representativity or diversity, indicating that both criteria are crucial for robust anomaly detection. Specifically, for the YelpHotel and C-15 datasets, there is a significant drop in performance without representativity. This can be attributed to the small scale of these datasets and is vulnerable to external noise. When incorporating external data without a similarity constraint, the model's understanding of normal patterns can become distorted, leading to decreased accuracy in anomaly detection. While for larger datasets like Twitter-20 and Tolokers, diversity plays a more critical role by enhancing the model's ability to capture a broader range of normal patterns.

\section{Regarding increasing the selection budget. }\label{app:budget}
We conducted experiments using 3 and 5 external data budgets within the Wild-GAD(OC/UL) settings, and we report the performance in Table~\ref{tab:budget1} and Table~\ref{tab:budget2}: Wild-GAD(OC/T/UL) and Wild-GAD(OC/F/UL), respectively. We have found that generally, after increasing the external data budget, the performance can have some improvement due to introducing more data and expose more normal patterns to the model, but the improvement can be very little and slightly, especially when the number of external data achieves 5. This can be explained by the fact that, after a certain point, additional external data may not contribute significantly to learning new normal patterns, as the model may have already captured the essential behaviors from a smaller subset of the external data.

\begin{table*}[!h]
\caption{Increasing selection budget AUC-PR Comparisons}
\vspace{-0.1in}
 \centering
 \resizebox{1.5\columnwidth}{!}{\begin{tabular}{lcccccc}
\toprule
{Method} & {YelpHotel} & {AmazonCN}  & {C-15}&{Twitter-20} & {Tolokers} & {Enron} \\ \hline
Wild-GAD(OC/S/UL) & 8.65(1.02)	 & 31.99(1.24) & 	86.85(3.32)	 & 62.32(1.12) & 	28.93(1.35)	 & 0.13(0.02)  \\ 
Wild-GAD(OC/M/UL)  & 	8.84(0.64)	 &35.38(0.54)	 &89.24(2.45)	 &64.23(1.25) &	30.21(1.21) &	0.14(0.03) \\ 
Wild-GAD(OC/T/UL)  & 8.95(1.34)& 	35.79(1.03)	& 88.49(2.12)	& 64.82(1.24)& 	30.06(1.26)& 	0.12(0.02)\\ Wild-GAD(OC/F/UL) &  8.82(1.25)& 	35.98(1.04)& 	89.13(1.97)& 	64.67(1.35)	& 29.67(1.44)& 	0.14(0.03)
\\ 
\bottomrule
\label{tab:budget1}
\end{tabular}}
\end{table*}

\begin{table*}[!h]
\caption{Increasing selection budget AUC-ROC Comparisons}
\vspace{-0.1in}
 \centering
 \resizebox{1.5\columnwidth}{!}{\begin{tabular}{lcccccc}
\toprule
{Method} & {YelpHotel} & {AmazonCN}  & {C-15}&{Twitter-20} & {Tolokers} & {Enron} \\ \hline
Wild-GAD(OC/S/UL) & 66.11(1.20)& 	80.03(2.57)	& 97.22(1.23)& 	54.72(2.32)& 	54.35(3.80)	& 66.08(3.10)  \\ 
Wild-GAD(OC/M/UL)  & 	68.55(1.22)&	81.43(1.23)	&97.72(1.78)	&57.68(1.35)&	60.19(0.45)&	64.12(0.25) \\ 
Wild-GAD(OC/T/UL)  & 68.29(1.08)&	82.19(1.29)&	97.49(1.36)	&58.02(1.22)&	59.98(2.26)	&66.55(1.01)\\ Wild-GAD(OC/F/UL) &  67.99(1.35)&	82.35(1.41)&	98.02(1.59)&	57.79(1.49)	&60.13(1.18)	&65.24(0.79)
\\ 
\bottomrule
\label{tab:budget2}
\end{tabular}}
\end{table*}
\section{GAE Backbone} \label{app:GAE}
Our framework is primarily designed with one-class SVDD as the backbone. However, we also seek to demonstrate the flexibility of our approach by incorporating GAE-based anomaly detection, which is a widely used and well-established method in the field. 
When employing GAE-based methods, the training objective and the score function have corresponding adaptations. GAE-based anomaly detection methods aim to leverage graph autoencoder to reconstruct the with the assumption that normal data can be reconstructed well while abnormalities will induce large reconstruction loss. Specifically, let $f_E$
 represent the graph encoder, 
 the graph decoder, and $Z = f_E(A, X) \in \mathbb{R}^{N \times d_h}$
 as the embeddings generated by the encoder. The training objective minimizes the reconstruction loss between the original adjacency matrix $A$
 and the reconstructed adjacency matrix $\hat{A} = \text{sigmoid}(ZZ^\top)$
, and between the original feature matrix $X$ and the reconstructed feature matrix 
$\hat{X} = f_D(A, Z)$. The training objective is the reconstruct loss, which is defined as follows: $\mathcal{L}_\text{GAE} = (1 - \alpha) \|A - \hat{A}\|_F^2 + \alpha \|X - \hat{X}\|_F^2$, 
 then we take the scoring function of the GAE-based model as the reconstruction loss itself to determine the normal nodes and anomalies.

 \section{Regarding relation with fine-tuning.}
Both our method and the conventional pre-train and fine-tune paradigm take advantage of external data, but there are notable differences between the two approaches.
The conventional graph pre-train and fine-tune paradigm leverages a large amount of external data during the pre-training phase to learn generalizable knowledge~\cite{when2023,Xu2023BetterWL,hafidi2020graphcl}. The pre-trained models are then fine-tuned on downstream tasks, using the specific target data to refine the learned features and align them more closely with the task requirements. In this framework, fine-tuning typically relies on the availability of a supervision signal from the downstream data, which guides the adjustment of the pre-trained model to better suit the target task~\cite{sun2024fine,huang2024measuring,fang2024universal}.
However, in our unsupervised anomaly detection, the goal is to learn the distribution of normal patterns from the target data. By training on the target graph first, the model captures the fundamental normal behaviors specific to that dataset. This ensures that the core representation learned by the model is closely aligned with the characteristics of the target data. Once the model has established a basic understanding of the target graph, we then introduce external graph data to expand its knowledge of normal patterns by incorporating diverse behaviors. The external data is used to supplement and reinforce the learned normal patterns, providing a richer context for the model.

\end{document}